\RequirePackage{fix-cm}
\documentclass[twocolumn]{svjour3}          % twocolumn
\smartqed  % flush right qed marks, e.g. at end of proof
\usepackage{graphicx}

\begin{document}

\title{HKR For Handwritten Kazakh \& Russian Database\thanks{This work was done with the support of grant funding for scientific projects of the MES RK No AR05135175 “Development and implementation of a system for recognizing handwritten addresses of written correspondence JSC “KazPost” using machine learning”.}
}


\author{
        Daniyar Nurseitov \textsuperscript{1,2}
        \and
        Kairat Bostanbekov \textsuperscript{1,2}       
        \and
        Daniyar Kurmankhojayev  \textsuperscript{3} %etc.
        \and
        Anel Alimova \textsuperscript{1,2} %etc.
        \and
        Abdelrahman Abdallah \textsuperscript{1,2} %etc.
}


\institute{
            Daniyar Nurseitov   \at
             nurseitovdb@gmail.com           %  \\
           \and
            Kairat Bostanbekov \at
            kairat.boss@gmail.com 
           \and
            Daniyar Kurmankhojayev \at
            kurman.daniyar@gmail.com
            \and
            Anel Alimova \at
            anic2002@mail.ru
            \and
            Abdelrahman Abdallah \at
            MSc Machine Learning \& Data Science \\
            Satbayev University\\
             abdoelsayed2016@gmail.com
             \and
             \textsuperscript{1} Satbayev University  Almaty, Kazakhstan 
             \textsuperscript{2} National Open Research Laboratory for Information and Space Technologies, Almaty, Kazakhstan
             \textsuperscript{3} Hong Kong Polytechnic University, Hung Hom, Hong Kong
}

\date{Received: date / Accepted: date}

\maketitle

\begin{abstract}
In this paper, we present a new Russian and Kazakh database (with about 95\% of Russian and 5\% of Kazakh words/sentences respectively) for offline handwriting recognition. A few pre-processing and segmentation procedures have been developed together with the database. The database is written in Cyrillic and shares the same 33 characters. Besides these characters, the Kazakh alphabet also contains 9 additional specific characters. This dataset is a collection of forms. The sources of all the forms in the datasets were generated by \LaTeX which subsequently was filled out by persons with their handwriting. The database consists of more than 1400 filled forms. There are approximately 63000 sentences, more than 715699 symbols produced by approximately 200 different writers. It can serve researchers in the field of handwriting recognition tasks by using deep and machine learning.

\end{abstract}

\section{Introduction}
\label{intro}
Today, handwriting recognition is a very urgent task. A solution to this problem would automate the business processes of many companies. One of the clear examples is a postal company, where the task of sorting a large volume of letters and parcels is an acute issue. 
Many researchers have made different types of handwritten text recognition systems for different languages such as English \cite{b2,b3,b4}, Chinese \cite{b5}, Arabic \cite{b9}, Japanese \cite{b6}, Bangla \cite{b7}, Malyalam \cite{b8}, etc. Having said that, the recognition problems of these scripts cannot be considered be entirely solved.

Any language contains a large number of words. For example, dictionaries of the Russian and Kazakh languages on average register more than 100,000 words, and the Oxford English dictionary more than 300,000 words. In this regard, collecting an exhaustive database of handwritten words, which include all words with a large variation in handwriting, seems almost impossible. In other words, there is always a word that the system cannot recognize.
\footnotesize
To the best of our knowledge, the analogs of handwritten text database for Russian and Kazakh languages do not exist.To create such a database, we decided to adopt the general principles of data collection and storage described in the IAM Database \cite{b1}. In the context of handwritten address recognition, it is necessary to identify the many keywords that can occur in the address. We utilized three different datasets described as following:
\begin{itemize}
    \item Handwritten samples (Forms) of keywords in Kazakh and Russian (Areas, Cities , Village , etc.)
	\item Handwritten Kazakh and Russian alphabet in cyrillic 
	\item Handwritten samples (Forms) of poems in Russian
\end{itemize}
\paragraph{In this paper, we describe the first version of a database that contains Russian words and also present a new database for offline handwriting recognition.}
The collection of this database combines the following steps. As an initial step, we collected the first data set with our own hands, since it is almost impossible to find such a set publicly available. 
This dataset was obtained by using forms, which consisted of machine-typed texts, and empty lines next to those texts. 
These empty lines were subsequently filled out by persons with their handwriting. 
It can serve as a basis for a variety of handwriting recognition tasks. Second, we collected handwritten Kazakh and Russian alphabet in Cyrillic. 
The last set of data came from handwritten samples of poems also filled by our own hands in Russian language.The databases were produced by approximately 200 different writers, each having 5 to 10 forms (made up of poem and keyword texts) to fill.  

For these purposes, we determined the minimum set of words, which includes all the names of cities, towns, villages, districts, and streets in Kazakhstan, and created layouts for filling out forms.
Forms were created in such a way as to simplify the process of “cutting” words from the form as much as possible (Fig. 1).
Extensive experiments related to the pre-processing of forms were also carried out in order to automatically identify forms, determine the contours of forms, compensate rotations, and also remove edge artifacts at the boundaries of segmented words.

\begin{figure}
  \includegraphics[width=\linewidth]{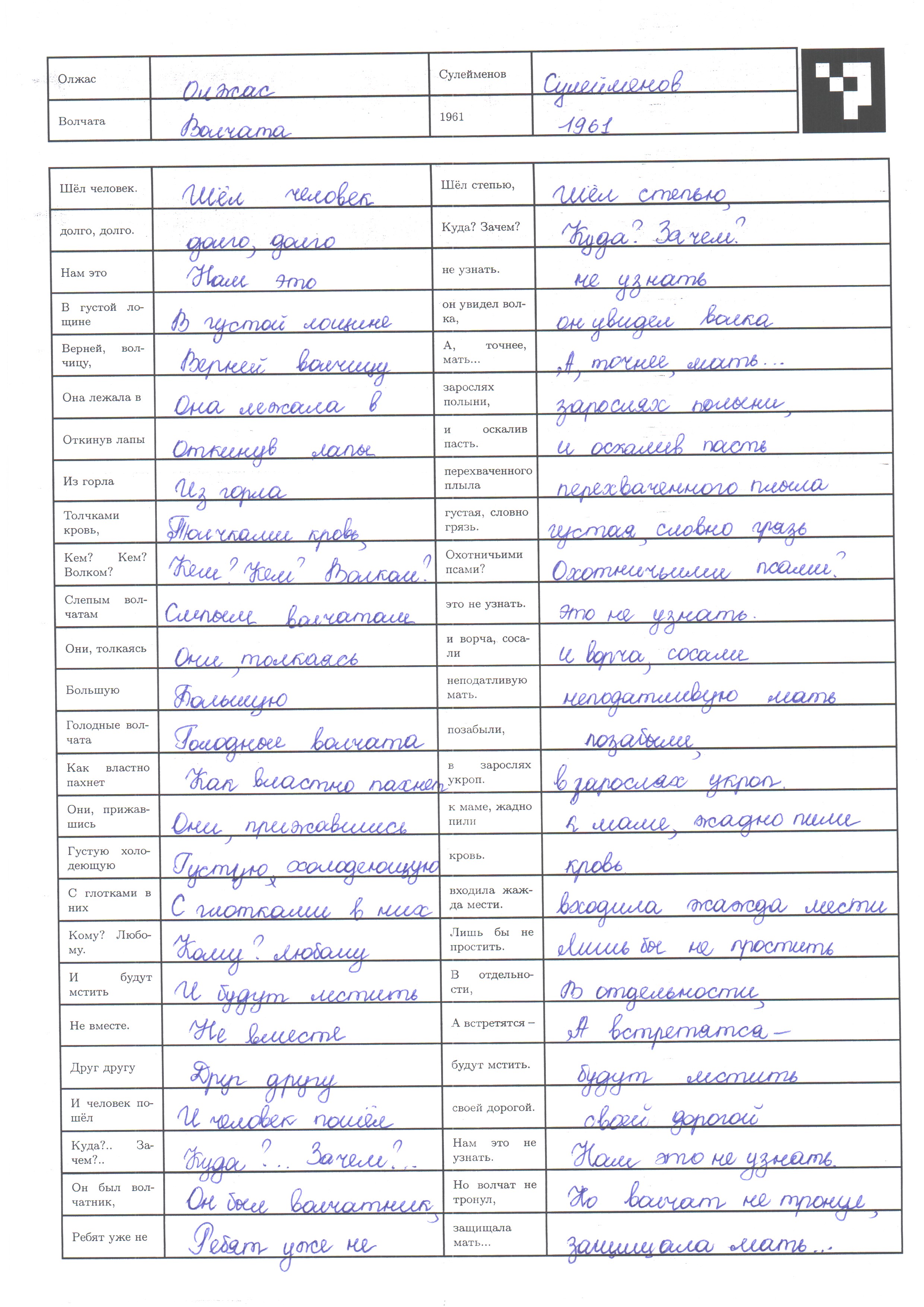}
\caption{One of the poem form in the dataset}
\label{fig:1}       % Give a unique label
\end{figure}
To solve the problem of recognition and processing of natural languages (natural language processing), which consists of optical recognition of characters of the manuscript texts in Russian and Kazakh languages, innovative software is being developed using state-of-the-art neural network-based machine learning methods.

The following section defines the related work on Handwriting Databases. Section 3 presents Data collection and storage phases is one of the most time consuming and costly stages. Section 4 provides Automated Labeling and Words Segmentation. Section 5 provides further characteristics of the Database, and concluding and future work are given in Section 6.

\section{Related Work}
The IAM Handwriting Database \cite{b1,b10} comprises handwritten samples in English which can be used to evaluate systems like text segmentation, handwriting recognition, writer identification and writer verification. The database is developed on the Lancaster-Oslo/Bergen Corpus and comprises forms where the contributors copied a given text in their natural unconstrained handwriting. Each form was subsequently scanned at 300 dpi and saved as gray level (8-bit) PNG image.The IAM Handwriting Database 3.0 includes contributions from 657 writers, making a total of 1539 handwritten pages comprising 5685 sentences, 13,353 text lines, and 115,320 words. The database is labeled at the sentence, line, and word levels. It has been widely used in word spotting \cite{b11,b12,b13,b14}, writer identification \cite{b15,b16,b17,b18,b19}, handwritten text segmentation \cite{b20,b21,b22}and offline handwriting recognition \cite{b23,b24,b25,b26}.

RIMES \cite{b27} is a representative database of an industrial application. The main idea of developing this database was to collect handwritten samples similar to those that are sent to different companies by postal mail and fax by individuals. Each contributor was assigned a fictitious identity and a maximum of up to five different scenarios from a set of nine themes. These themes included real-world scenarios like ‘damage declaration’ or ‘modification of contract’. The subjects were required to compose a letter for a given scenario using their own words and layout on a white paper using black ink. A total of 1300 volunteers contributed to data collection, providing 12,723 pages corresponding to 5605 mails. Each mail contains two to three pages, including the letter written by the contributor, a form with information about the letter, and an optional fax sheet. The pages were scanned, and the complete database was annotated to support evaluation of tasks like document layout analysis \cite{b28}, mail classification \cite{b29}, handwriting recognition \cite{b30} and writer recognition \cite{b17}.
\footnotesize
The National Institute of Standards and Technology, NIST, developed a series of databases \cite{b31} of handwritten characters and digits supporting tasks like isolation of fields, detection and removal of boxes in forms, character segmentation, and recognition.The form comprises boxes containing writer information, 28 boxes for numbers and 2 for alphabets, and 1 box for a paragraph of text. The NIST Special Database 1 comprised samples contributed by 2100 writers. The latest version of the database, the Special Database 19, comprises handwritten forms of 3600 writers with 810,000 isolated character images along with ground truth information. This database has been widely employed in a variety of handwritten digits \cite{b32} and character recognition systems \cite{b33}.

CVL \cite{b34} is a database of handwritten samples supporting handwriting recognition, word spotting, and writer recognition. The database consists of seven different handwritten texts, one in German and six in English. A total of 310 volunteers contributed to data collection, with 27 authors producing 7 and 283 writers providing 5 pages each. The ground truth data is available in XML format, which includes transcription of text, the bounding box of each word, and the identity of the writer. The database has been used for writer recognition and retrieval \cite{b35} and can also be employed for other recognition tasks. In addition to regular text, a database of handwritten digit strings written by 303 students has also been compiled \cite{b36}. Each writer provided 26 different digit strings of different lengths, making a total of 7800 samples. Isolated digits were extracted from the database to form a separate dataset—the CVL Single Digit Dataset. The Single Digit Dataset comprises 3578 samples for each of the digit class (0-9). A subset of this database has also been used in the ICDAR 2013 digit recognition competition \cite{b36}.

The AHDB \cite{b37,b38} is an offline database of Arabic handwriting together with several pre-processing procedures. It contains Arabic handwritten paragraphs and words. Words used to represent numbers on checks produced by 100 different writers. The database was mainly intended to support automatic processing of bank checks, but it also contains pages of unconstrained texts allowing evaluation of generic Arabic handwriting recognition systems as well. The database was employed in handwriting recognition \cite{b39} and writer identification tasks \cite{b40}.

\section{Data collection and storage}

\subsection{Data collection}
A data collection phase is one of the most time consuming and costly stages. Our main task is to simplify and automate as much as possible. 
The sources of all the forms in the datasets were generated by \LaTeX, then converted to PDF and printed to be filled by writers. So, it was an easy  task  to  generate  the correct labels for the printed text on the forms.Each writer filled approximately between 5-10 forms from keyword and poem forms,so each form in the dataset is written approximately between 5-10 writers.Each form has a unique id at the name of the form.. The word or letter is placed in the rectangle.The filled forms and letters were been scanned with a Canon MF4400 Series UFRII scanner at a resolution of 300 dpi and a color depth of 24 bits.

We collected three different Datasets described as the following:
    \begin{itemize}
	%\item scans of the front side of envelopes shown in Fig. 3.
	\item Handwritten samples (Forms) of keywords in Kazakh and Russian (Areas, Cities , Village , etc.) are shown in Fig. 2.
	\item Handwritten Kazakh and Russian alphabet in Cyrillic are shown in Fig. 2.
	\item Handwritten samples (Forms) of poems in Russian are shown in Fig. 1.
    \end{itemize}
The next step was to annotate the collected data, i.e. to mark and synthesize new samples from existing ones, using various geometric and photo-metric transformations (data augmentation).

\begin{figure}[!ht]
  \centering
  %\begin{subfigure}[b]{0.6\linewidth}
    \includegraphics[width=0.6\linewidth]{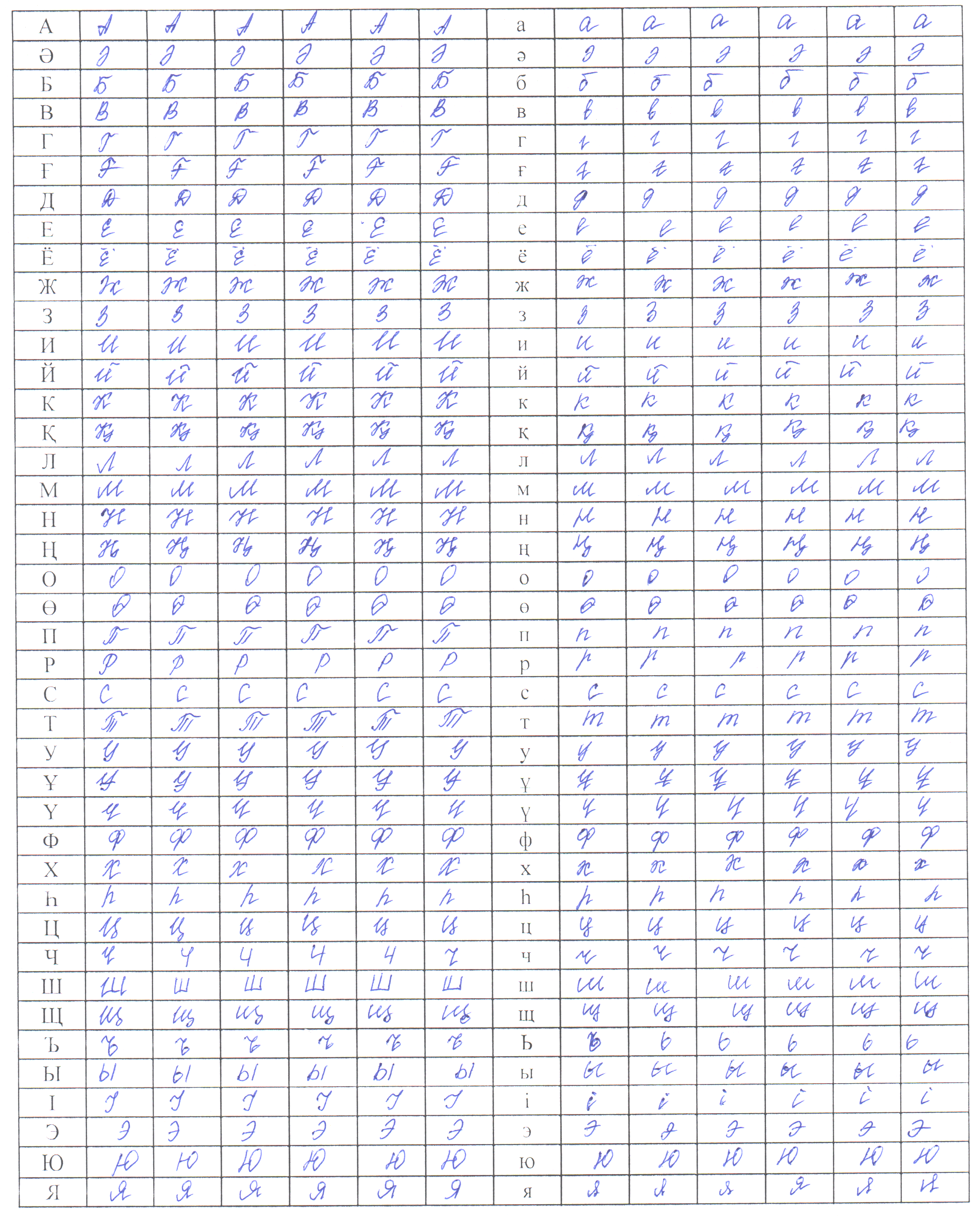}
  %\end{subfigure}
  %\begin{subfigure}[b]{0.6\linewidth}
    \includegraphics[width=0.6\linewidth]{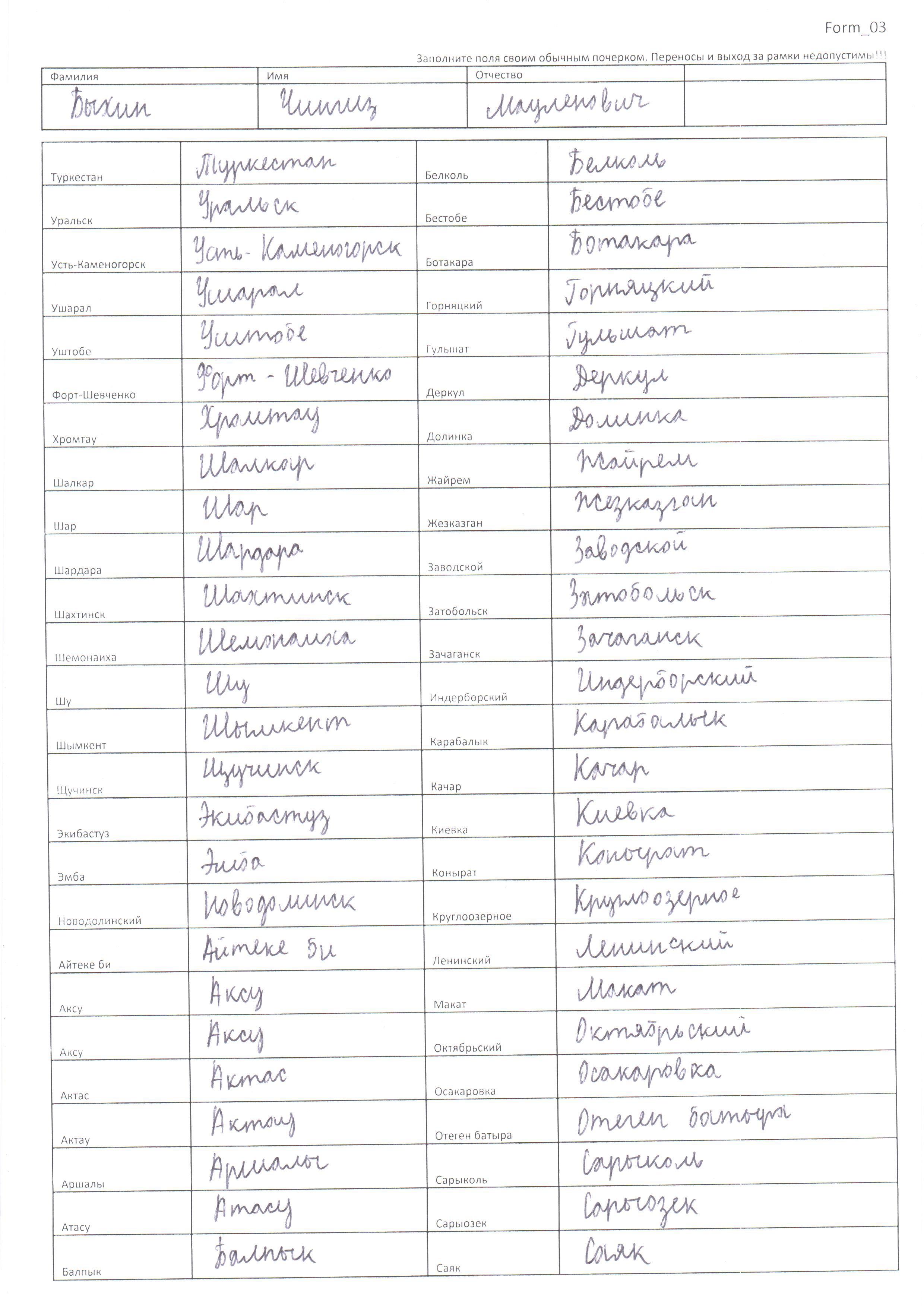}
  %\end{subfigure}
  \caption{Two forms for collecting handwritten samples of the Cyrillic alphabet and Keywords.}
\end{figure}

\subsubsection{Keyword Database}
To begin with, we consider correspondence addresses relevant for the Republic of Kazakhstan, as the list of keywords containing the following names:
\begin{itemize}
\item Areas
\item Cities
\item Village
\item Settlements
\item Streets
\item Poems 
\item Russian Letter 
\end{itemize}

Additional information, such as:
\begin{itemize}
\item Indices
\item Phones
\item Surnames
\item Company Names
\end{itemize}
were not included in the database.
\subsubsection{Handwritten Alphabet and Forms}
There are two fundamental approaches to text recognition: character recognition (Optical Character Recognition, OCR) and word recognition (Optical Word Recognition, OWR). With OCR, a model dataset required to train the model should contain handwritten samples of all the characters in a language alphabet. It is an important for each language to compose separate forms, since the set of letters of different alphabets can vary greatly. On the other hand, with OWR,  a model dataset required to train the model should contain handwritten samples of all the Words for the language. Further, for subsequent training and testing of the model, handwritten samples of target words are needed. An example of one of the forms for collecting word samples and letters (Fig. 2).

\subsubsection{Data Collection Methods}
A person who has agreed to provide a sample of his handwriting will fill the forms and give the form to us and we scan and save it in our database.
\section{Automated Labeling and Words Segmentation}
\subsection{Automated Labeling}
Labeled data are data that have been marked with labels identifying certain features, characteristics, or a kind of object. The labeling of data is a prerequisite for recognition experiments. Labeling data is expensive, time consuming, and error prone."like in IAM Dataset" \cite{b1}, we decided to do as much automation as possible automatically. The sources of all the forms printed (and subsequently filled by writers) were saved on a text file with a unique id for the form and the cell number in the form. So it was an easy task to generate the correct labels for the printed text on the forms.
In this regard, we have developed a recommendation system that allows us to simplify the process of labeling data in forms.
\subsection{Segmentation}
The form is designed so that it is possible to easily identify and segment by cells. To identify the form, there is a marker in the upper-right corner of each form. To simplify the process of segmentation, the entire form is divided by horizontal and vertical lines, which makes it quite easy to restore the structure of the document, and accordingly, the spatial position of the word. Words are indexed (annotated) according to their position in the table.
In order to cut out cells from the form, the following actions (pre-processing) are performed:
\begin{itemize}
\item filtering forms to enhance table boundaries
\item defining of the contours of the table
\item Determination and compensation of the angle of rotation
\item exclusion of lines
\item Sorting forms by id (marker)
\item Streaming the division of forms into words
\item name and storage of words
\end{itemize}
After the image areas corresponding to the word cells are segmented, segmented image areas corresponding to the word cells may contain some edge artifacts. For example, line artifacts cut out with a cell or parts of a word from a neighboring cell can be attributed to artifacts (Fig. 3). We eliminate these artifacts by constructing vertical and horizontal histograms (Fig. 4 , Fig. 5) also by cutting off parts that are separately localized closer to the edges of the cell.

\begin{figure}[!ht]
  \centering
  %\begin{subfigure}[b]{0.6\linewidth}
    \includegraphics[width=0.6\linewidth]{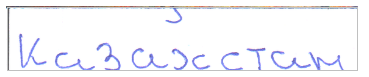}
  %\end{subfigure}
  %\begin{subfigure}[b]{0.6\linewidth}
    \includegraphics[width=0.6\linewidth]{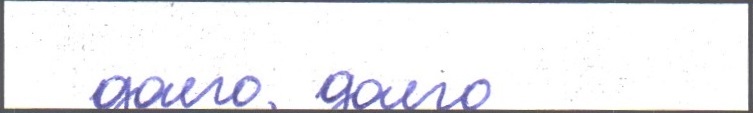}
  %\end{subfigure}
  \caption{Example of a region cut out of a form with a word. A pronounced cell line and a piece of letter from a neighboring area are visible along the edges.}
  \end{figure}

However, it is not always possible to eliminate all artifacts. The following are some aspects that make further processing of a segmented word difficult:
\begin{itemize}
\item Letters may not be interconnected.
\item Letters can be perfected with artifacts.
\item The position of the letters and their size vary significantly from word to word.
\item Letters can be written in different colors (blue, black, red).
\end{itemize}
In this regard, we have developed a recommendation system that allows us to simplify the process of selecting areas with words from the form.
\begin{itemize}
\item We suggest filling out the form using the blue pen. This will allow the system to distinguish the word from the table borders at the color level. For example, by converting an image from RGB to HSV, we get a color representation of objects that is invariant with respect to lighting. In this color space, blue remains blue, regardless of the brightness and intensity of the image.

\item sometimes eliminating parts of words from neighboring cells is impossible without distorting the target content of a given cell; therefore, when filling out the form, it is desirable that the subject does not go beyond the boundaries of the cell.

\item find the region of interest (ROI) in forms ,the ROI in our forms is two columns that are filled by writers.  

\item We segmented the cells depending on the horizontal white space between the cell by using the histogram (Fig. 4).

\item then we exclusion of lines that cross the words.

\item Finally, we cropped and segmented the cells depending on the vertical white-space by using the histogram (Fig. 5).
\end{itemize}

\begin{figure}
     \centering
        \raisebox{-\height}{\includegraphics[width=0.40\textwidth]{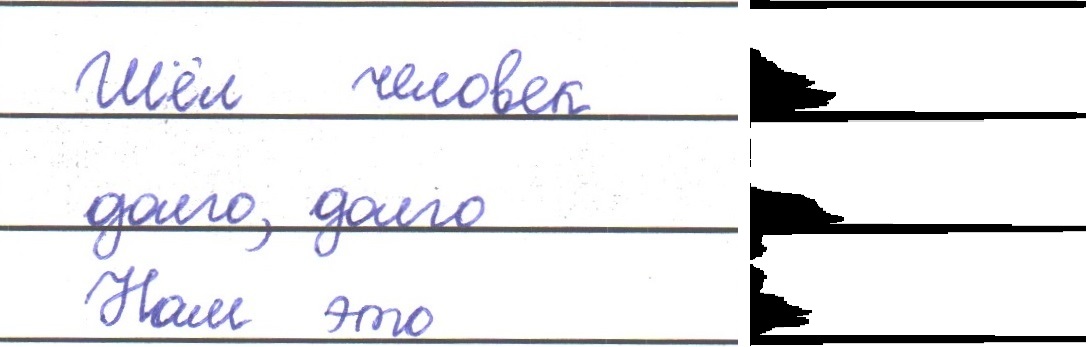}}
        \caption{Horizontal histogram}
   \end{figure}    
    \hfill
    %\begin{figure}
    %    \raisebox{-\height}{\includegraphics[width=0.40\textwidth]{step02_noline.jpg}}
    %    \caption{Example 2}
    % \end{figure}
    \begin{figure}
    \centering
        \raisebox{-\height}{\includegraphics[width=0.40\textwidth]{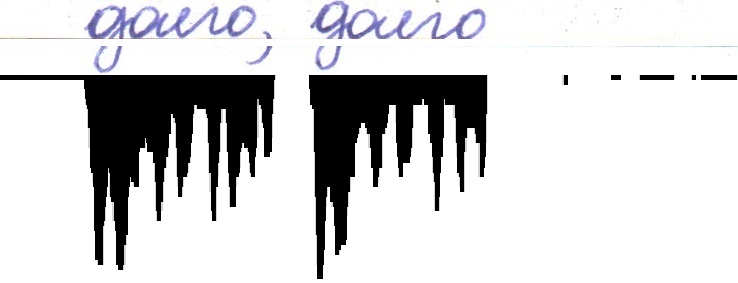}}
        \caption{Vertical histogram}
    \end{figure}
    \begin{figure}
    \centering
        \raisebox{-\height}{\includegraphics[width=0.40\textwidth]{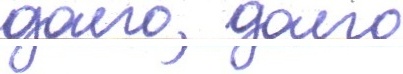}}
        \caption{Examples of segmented words.}
\end{figure}

Next, we normalize the words images, as follow :
\begin{itemize}
\item cast all word images to the same size
\item center the word on the image
\item reduce image size
\end{itemize}
The final image shown in Fig. 6.
\section{Further Characteristics of the Database}

The database consists of more than 1400 filled forms written by 200 writers. There are approximately 63000 sentences, more than 715699 symbols shown in Fig. 7. And also There are approximately 106718 words.
total images in the dataset after pre-processing and segmentation the forms are 64943 images.

\section{Conclusion and Future Work}

We have built the handwritten Kazakh, Russian database. The database can serve as a basis for research in handwriting recognition.
This contains Russian Words (Areas, Cities, Village, Settlements, Areas, Streets) by a hundred different writers.
It also incorporates the most popular words in the Republic of Kazakhstan.A few pre-processing and segmentation procedures have been developed together with the database.
Finally, it contains free handwriting forms in any area of the writer interest. This database is meant to provide a training and testing set for Kazakh, Russian Words recognition research.In future, as further work on gathering Handwriting samples of keywords and envelope shots will continue.At the same time, envelopes are annotated and various metrics are checked to evaluate the recognition error.In order for the artifacts to not interfere, we need to collect as much tagged data as possible.

\begin{figure}[!ht]
  \centering
 
  %\begin{subfigure}[b]{\linewidth}
    \includegraphics[width=\linewidth]{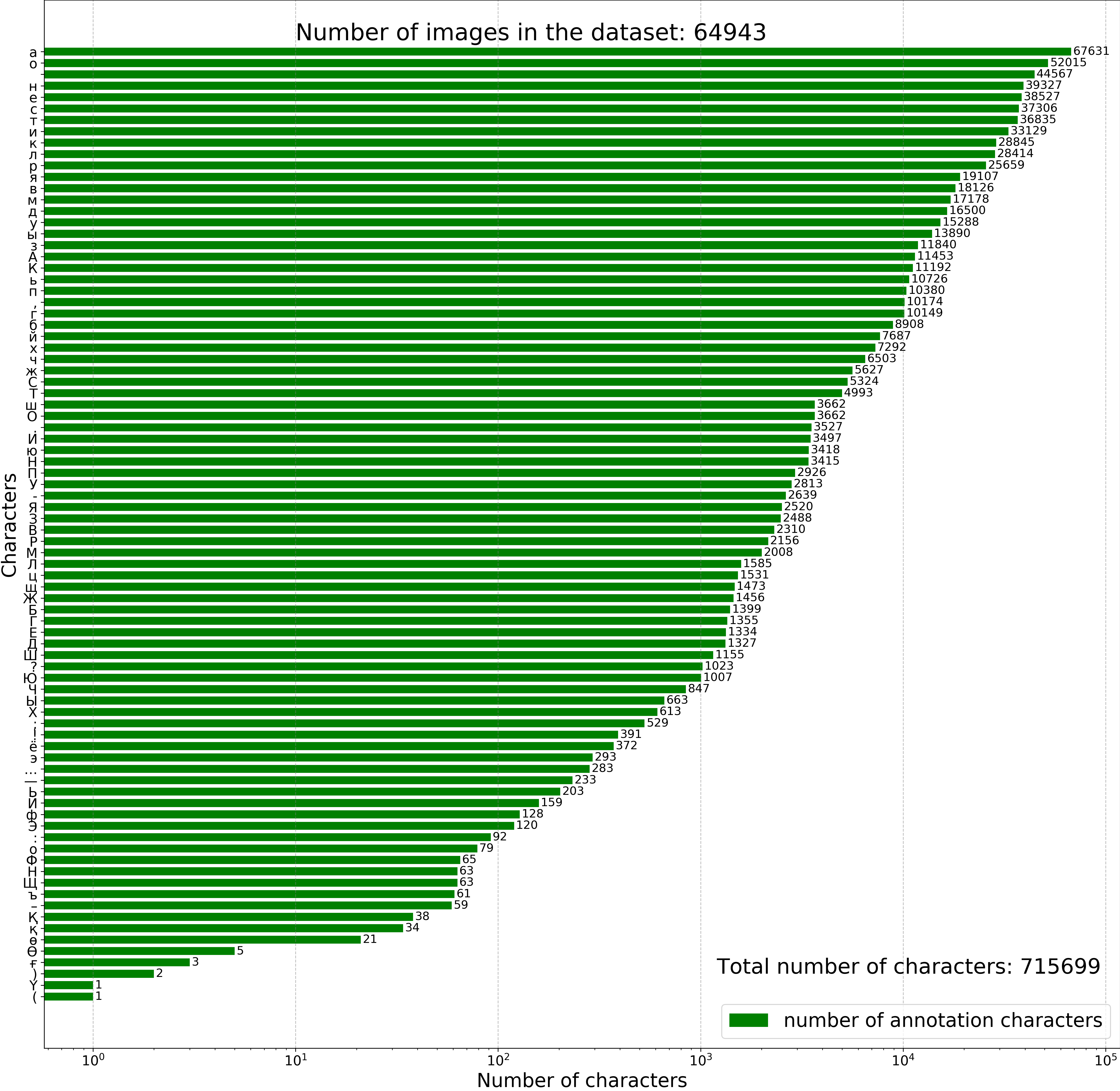}
    
  %\end{subfigure}
  
  \caption{Histogram of Characters in the dataset}
  
\end{figure}

\begin{acknowledgements}
This work was funded by the Ministry of Education and Science of the Republic of Kazakhstan (Grant No AP05135175)
\end{acknowledgements}




\end{document}